\documentclass[lettersize,journal]{IEEEtran}
\usepackage{amsmath,amsfonts}
\usepackage{amssymb}
\usepackage{booktabs}
\usepackage{makecell}
\usepackage{algorithmic}
\usepackage{algorithm}
\usepackage{array}
\usepackage[caption=false]{subfig}
\usepackage{textcomp}
\usepackage{stfloats}
\usepackage{url}
\usepackage{verbatim}
\usepackage{graphicx}
\usepackage{cite}
\usepackage[colorlinks=true, linkcolor=blue, citecolor=green, urlcolor=red]{hyperref}
\usepackage{multirow}

\begin{document}

\title{NavIsaacLab: Generating Realistic Crowd via Parallel Robot Learning for Benchmarking\\Human-aware Navigation}

\author{
Anonymous Submission
}
\author{
    Bingyi~Xia$^{\dag}$, 
    Han~Bao$^{\dag}$, 
    Jingyu~Zhu, 
    Hanjing~Ye, 
    Yuhan~Pang,
    Guangcheng~Chen,\\ 
    Liang~Lin, 
    Wenjun~Xu and
    Jiankun~Wang 
\thanks{$^{\dag}$ Equal contribution.}
\thanks{Corresponding authors: Wenjun~Xu, Jiankun Wang}
\thanks{Bingyi Xia, Han Bao, Hanjing Ye, Yuhan Pang, Guangcheng Chen are with Department of Electronic and Electrical Engineering, Southern University of Science and Technology, Shenzhen, China.}
\thanks{Jingyu Zhu is with National University of Singapore, Singapore, Singapore}
\thanks{Liang Lin, Wenjun Xu are with the Research Institute of MA\&EI, Peng Cheng Laboratory, Shenzhen, China.{\tt \small (xuwenjunwendy@gmail.com)}} 
\thanks{Jiankun Wang is with the Shenzhen Key Laboratory of Robotics Perception and Intelligence, Department of Electronic and Electrical Engineering, Southern University of Science and Technology, Shenzhen, China, and also with the Jiaxing Research Institute, Southern University of Science and Technology, Jiaxing, China.{\tt \small(wangjk@sustech.edu.cn)}}
}

\markboth{Journal of \LaTeX\ Class Files,~Vol.~14, No.~8, August~2021}%
{Shell \MakeLowercase{\textit{et al.}}: A Sample Article Using IEEEtran.cls for IEEE Journals}


\maketitle

\begin{abstract}
Robot autonomous navigation that accounts for surrounding human activities is crucial for ensuring both safety and natural human-robot interaction in real-world environments shared by humans and robots.
Simulation of complex and diverse navigation scenarios serves as the foundation for training reliable robot navigation policies and accurately evaluating the performance of algorithms, offering an efficient alternative to manual supervision of real data. 
However, current human-aware navigation research faces significant challenges due to the scarcity of diverse, high-quality scene data. 
Existing simulation platforms often rely on handcrafted rules to approximate pedestrian behavior and lack the capability to provide extensive sensor signals, typically assuming perfect observations.
To address these limitations, this paper presents NavIsaacLab, a comprehensive framework for benchmarking and training human-aware navigation policies through physics-based and photo-realistic simulations of pedestrians and scenes.
Based on Isaac Lab, the proposed framework employs photo-realistic scene rendering capabilities and supports parallel simulation on GPU, delivering real-time and accurate 3D visual feedback to robots.
To enhance the realism of human behavior, a data-driven approach is employed that incorporates a trajectory diffusion model and an adversarial motion learning controller, enabling controllable, physics-based pedestrian simulation. 
Furthermore, the integration of diverse cross-scale scenes provides a robust benchmark for state-of-the-art human-aware navigation methods.
\end{abstract}



\section{Introduction}

Robots are increasingly expected to operate in complex and dynamic environments shared with humans, where ensuring safety and natural human-robot interaction (HRI) are paramount.
Such robotic applications with inevitable human presence, including delivery robots and assistive robots\cite{YUEN2022Acceptance, xia2023Collaborative, Hwang2023system} have highlighted the necessity of human-aware navigation as a fundamental functionality.
Unlike traditional navigation approaches that primarily focus on obstacle avoidance and path efficiency, human-aware navigation also prioritizes human comfort and expectations. 
This relies on decision-making grounded in comprehensive situational awareness, which encompasses environmental context, human interactions (both human-human and human-robot), and social norms\cite{singamaneni2024survey}.


\begin{figure}[!t]
\centering
    \includegraphics[width=0.98\linewidth]{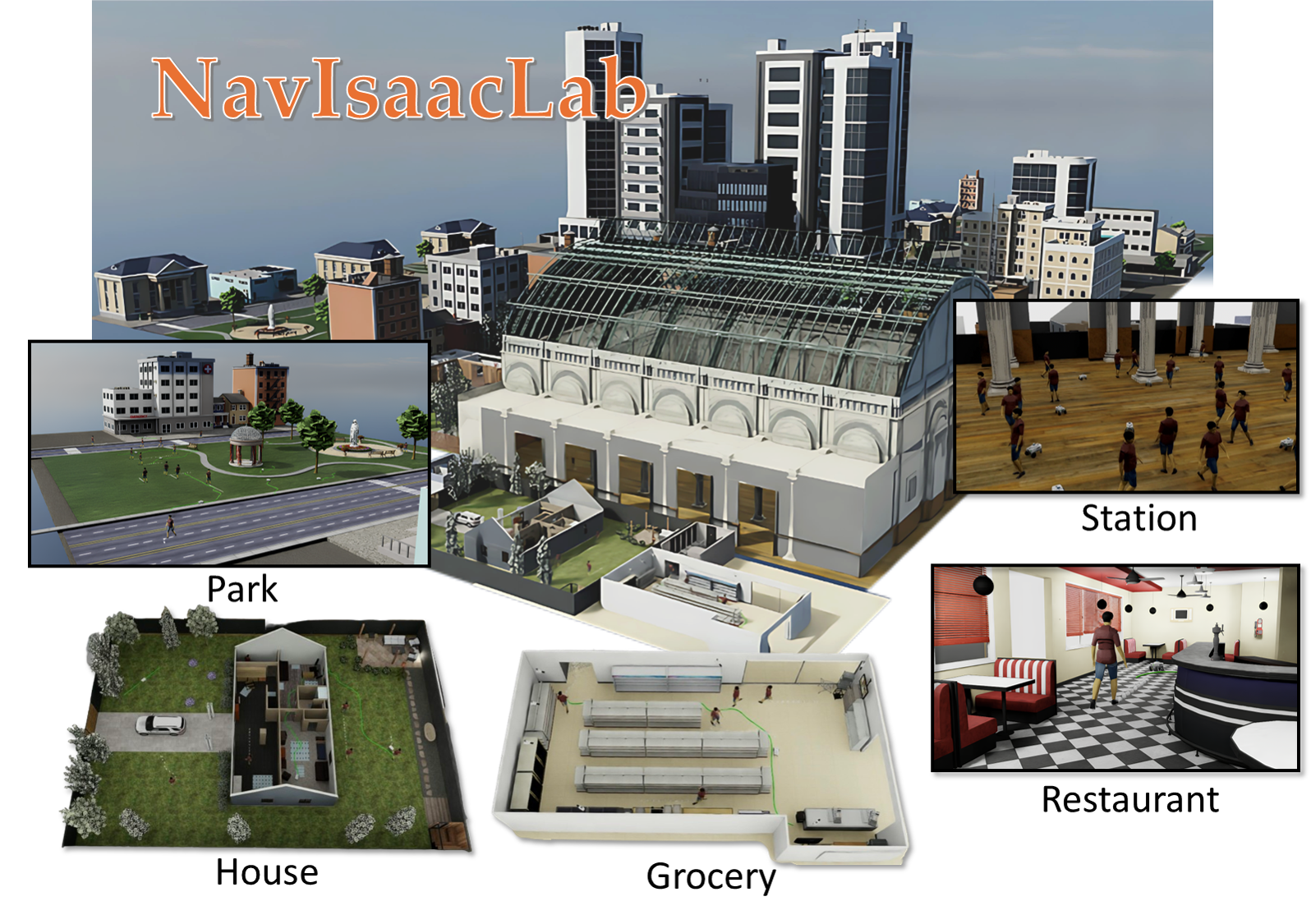}
    \caption{
    NavIsaacLab offers realistic and diverse scenarios required for training and evaluating human-aware navigation. Based on the framework of Isaac Lab, our platform features realistic rendering of the scenes, physically controllable pedestrian simulation, and parallel training capability.
    }
    \label{fig:demo}

\end{figure}

A critical challenge in developing human-aware navigation systems is the need for extensive HRI data in training and testing.
In practice, simulation of human reactions to robot behavior has been widely adopted as an alternative to real-world datasets, as it can offer flexible environment adjustments, eliminate the need for manual annotation, and support risk-free exploration\cite{Francis2024Principles}.
Current simulation data generation and collection processes require further improvement, rather than simplifying pedestrian behavior by reducing it to planar position and velocity representations.
On the one hand, the vision-based understanding of human behavior relies on generating diverse multimodal data, as these 3D behavioral cues inherently shape human intentions.
Recent studies have incorporated implicit signals (e.g., gaze\cite{chen2020gaze}, body orientation\cite{Zhao2024human, Wong2024SocialCircle}, and gesture\cite{Salzmann2023robotscansee, saadatnejad2024socialtransmotion}) to further enhance bidirectional understanding in shared environments.
On the other hand, parallel simulation significantly enhances the scalability of data collection, particularly for reinforcement learning (RL) based human-aware navigation.
For instance, \cite{cai2021hyp, Matsuzaki2022learning} have demonstrated that parallel data processing in RL significantly speeds up training and enhances navigation performance.
To support the development of new paradigms in human-aware navigation, it is crucial to establish a standardized benchmark compatible with diverse methods that can provide fair and reliable algorithm evaluation. 
However, a significant gap remains in the fidelity of the existing simulation benchmarks and their real-world applicability. 
We summarize that three critical issues persist:
\textbf{1) Simulation realism:} Since one of the advantages of simulation is the support for extensive sensor and ground truth, a simulation benchmark should enable photorealistic rendering and simulate physical dynamics, rather than simplified 2D trajectory approximations.
\textbf{2) Pedestrian behavior fidelity:} Most simulation benchmarks\cite{tsoi_sean_2022, higueras_hunavsim_2023, stratton2024complexity} rely on limited handcrafted interactions and pre-recorded procedural animations. 
Although spatial relationships between agents can be observed, the multi-modality of interaction behaviors is compromised.
\textbf{3) Computational scalability:} Advances in physics engines enable parallel simulation, allowing simultaneous training of multiple agents in diverse scenarios.
This capability substantially increases sample throughput, thereby accelerating policy training and evaluation, while also promoting cross-domain generalization.
To address these limitations, we propose a comprehensive benchmark framework NavIsaacLab to bridge the gap between simulated and real-world environments, offering researchers a powerful tool to develop, test, and refine their human-aware navigation policies.
This framework integrates physics-based and photorealistic simulations of both pedestrians and scenes, thereby providing a more accurate and dynamic representation of real-world conditions. 
By leveraging the advanced capabilities of NVIDIA Isaac Lab\cite{mittal2025isaaclab}, our system supports GPU-based parallel simulation, which together delivers real-time and accurate 3D visual feedback to robotic systems.
Besides, we enhance the fidelity of human behavior modeling by employing a data-driven approach\cite{rempeluo2023tracepace}. 
This approach integrates a trajectory diffusion model with an adversarial motion learning controller to enable controllable, physics-based pedestrian simulation. 
Such integration not only improves the realism of simulated pedestrian movements but also offers a robust platform for testing and benchmarking state-of-the-art human-aware navigation methods across diverse, cross-scale scenarios.
In addition to establishing a high-fidelity simulation benchmark, we further propose a human-aware navigation policy that leverages ego-centric visual cues of the pedestrian motion provided by NavIsaacLab. 
Our policy incorporates pedestrian body orientation to infer motion intention through a Transformer encoder, and integrates this enriched representation into a DRL navigation framework.
This demonstrates NavIsaacLab’s capability to support vision-based policy learning for navigation in crowds.
The main contributions are threefold:
\begin{itemize}
    \item \textbf{Simulation Platform:} We introduce NavIsaacLab, a novel simulation benchmark for human-aware navigation tasks built on the Isaac Lab platform, which significantly improves simulation efficiency while providing rich visual feedback. 
    It supports GPU-based parallel simulation, enabling scalable data generation and efficient training of learning-based navigation policies.
    Furthermore, it incorporates a high-fidelity pedestrian behavior authoring method, allowing pedestrians to exhibit realistic movement through joint-level motion control.

    \item \textbf{Benchmark Development:} We establish a comprehensive benchmarking framework for evaluating human-aware navigation algorithms. 
    The benchmark provides a multi-dimensional metric system to assess robot performance in simulated human-populated environments.
    Through systematic evaluations across 30 photorealistic scenes with diverse layouts, we demonstrate the platform’s capability to deliver detailed analyses, including studies on human density impacts and human-robot avoidance dynamics in crowd-aware environments.

    \item \textbf{Baseline Algorithm:} We propose a reinforcement learning navigation strategy incorporating human pose perception inputs. 
    The approach introduces a posture-corrected human-robot interaction attention mechanism to model pedestrian movement intention. 
    We demonstrate the platform's utility for synthetic data collection, rapid policy training, and successful simulation-to-reality transfer through physical robot experiments, validating the practical effectiveness of our methodology.

\end{itemize}


\section{Related Work}

\subsection{Human-aware Navigation}

\begin{table*}[ht]
\centering
\caption{Comparison with Other Existing Benchmarks}
\label{tab:related}
\resizebox{0.98\textwidth}{!}{%
\begin{tabular}{ccc ccc cc}

    \toprule
    \textbf{Benchmark} & \textbf{Year} & \makecell{\textbf{Simulation}\\ \textbf{Platform}} & \makecell{\textbf{Parallel}\\ \textbf{Simulation}} & \makecell{\textbf{Physical}\\ \textbf{Fidelity}} & \makecell{\textbf{Pedestrian}\\ \textbf{Modeling}} & \makecell{\textbf{Pedestrian Behavior}\\ \textbf{Authoring Method}} & \textbf{Visual Fidelity}\\
    
    \midrule
    \makecell{gym-avoid\\ \cite{everett2018motion}} & 2021 & - & - & Kinematics & Trajectory & ORCA & 2D Geometry \\[6pt]
    
    \makecell{SocNavBench\\ \cite{biswas_socnavbench_2022}} & 2022 & - & - & Kinematics & Trajectory & Trajectory Replay & 2D Geometry \\[6pt]
    
    \makecell{SEAN 2.0\\ \cite{tsoi_sean_2022}} & 2022 & Unity/Gazebo & $\times$ & Dynamics & Scripted Animation & SFM & Photorealistic Model \\[6pt]
    
     \makecell{HuNavSim\\ \cite{higueras_hunavsim_2023}} & 2023 & Gazebo & $\times$ & Kinematics & Scripted Animation & SFM & 3D Model \\ [6pt]

    \makecell{HabiCrowd\\ \cite{vuong2024habicrowd}} & 2024 & Habitat & $\times$ & Kinematics & Scripted Animation & UMP model & Reconstruction Mesh  \\[6pt]
    
    \makecell{ComplexityNav\\ \cite{stratton2024complexity}} & 2025 & - & - & Kinematics & Trajectory & SFM / ORCA & 2D Geometry \\[6pt]
    
    \makecell{Arena 5.0\\ \cite{kastner_demonstrating_2024}} & 2025 & Unity/Gazebo/Isaac Sim & $\times$ & Dynamics & Scripted Animation & SFM / ORCA & Photorealistic Model  \\[6pt]
    
    \midrule
    \makecell{\textbf{NavIsaacLab}\\ \textbf{(Ours)}} & 2025 & Isaac Lab & \checkmark & Dynamics & Joint-controlled Motion & Diffusion+AMP & Photorealistic Model  \\ 
\bottomrule
\end{tabular}
}
\end{table*}

Human-aware navigation has been studied for decades, which has evolved from basic collision avoidance to minimizing disturbance and discomfort to surrounding humans.
Early research primarily focused on safety, where reaction-based strategies\cite{Snape2011hrvo} and RL-based methods\cite{chen2017sarl, Chen2019crowd} were introduced to enable collision avoidance based on 2D geometry and velocity observations.
In recent years, beyond safety and efficiency, researchers have increasingly focused on human comfort, naturalness, and sociability to further adapt to highly crowded and complex dynamic environments. 
Improvements in these areas are anticipated by incorporating higher-level social norms, including human intention\cite{liu2023intension}, proxemics\cite{Medina2023Proxemic}, and group formations\cite{cai2023sampling, Luo2025GSON}.
Furthermore, gaits, gestures, and gaze direction are explored as implicit cues\cite{chen2020gaze, Zhao2024human, Salzmann2023robotscansee, Wong2024SocialCircle} for precisely predicting human intention, as they provide contextual information about human interactions. 
More recently, large-scale foundation model approaches, such as end-to-end vision-language-action models\cite{Song2025VLM}, have demonstrated strong scene understanding and generalization across diverse scenarios by exploiting rich multimodal sensory data.
However, evaluating these methods remains challenging due to the lack of standardized benchmarks that incorporate pedestrian visual signals and full-body motion data.

\subsection{Simulation Benchmarks}

Simulation benchmarks have been widely adopted to provide standardized evaluation metrics and quantify social navigation principles for the development of human-aware navigation methods. 
The early platform gym-avoidance \cite{everett2018motion} established foundational frameworks by providing rule-based pedestrian behavior models, geometric collision feedback, and metrics such as collision rates, while supporting reinforcement learning training.
Subsequent works expanded functionality across three dimensions: scenario synthesis, pedestrian behavior modeling, and benchmark protocols.
ComplexityNav\cite{stratton2024complexity} proposed scenario complexity that classified congestion levels and passageway topologies to instruct scenario synthesis.
Arena 5.0\cite{kastner_demonstrating_2024} standardizes evaluation protocols that facilitate cross-platform algorithm comparisons. 
SocNavBench\cite{biswas_socnavbench_2022} advanced pedestrian behavior visualization by enabling perspective rendering of trajectory datasets that extend behavioral modalities beyond pure trajectory representation.

Physics-based simulation benchmarks have emerged to bridge the sim-to-real gap.
For example, HuNavSim\cite{higueras_hunavsim_2023} integrates Gazebo to simulate pedestrian interactions with environmental contexts, while SEAN 2.0\cite{tsoi_sean_2022} leverages Unity for higher fidelity rendering.
Although they provide adjustable virtual scenes with handcrafted HRI designs, the adaptability of evaluation results remains constrained by the small number of manually configured environments and the few human-robot interaction scenarios.
HabiCrowd\cite{vuong2024habicrowd} embeds diverse pedestrian models into 3D indoor reconstruction scenes via Habitat, constructing a large-scale dynamic scenario dataset for human-aware visual navigation.

Despite these advancements, critical challenges persist: existing benchmarks often oversimplify pedestrian behaviors without contextual visual signals and lack the protocol of parallel simulation for RL-based methods.
NavIsaacLab aims to construct a simulation benchmark that provides these features to facilitate fair comparisons and support future development.
A detailed comparative analysis of these benchmarks is summarized in Table \ref{tab:related}.
Based on Isaac Lab\cite{mittal2025isaaclab}, our platform enables fast and flexible photorealistic rendering and high-fidelity physics, improving realism compared to prior benchmarks~\cite{biswas_socnavbench_2022, tsoi_sean_2022, stratton2024complexity}.
It enables parallel simulation through GPU-based parallelization that efficiently scales to a large number of environments. 
This distinguishes NavIsaacLab from prior benchmarks and improves sample efficiency and training scalability.
Furthermore, by introducing a novel behavior authoring method, NavIsaacLab supports fully controllable pedestrian motion with realistic physical interactions, allowing more fine-grained and expressive pedestrian modeling than trajectory or animation-based approaches\cite{kastner_demonstrating_2024, vuong2024habicrowd}.

\subsection{Pedestrian Behavior Authoring Methods}

Pedestrian behavior authoring in simulations serves two critical functions for human-aware navigation: reactive movement and whole-body motion such as walking gaits and postures. 
For reactive movement generation, most benchmarks\cite{tsoi_sean_2022, higueras_hunavsim_2023, stratton2024complexity, vuong2024habicrowd, kastner_demonstrating_2024} employ rule-based models for pedestrian behavior like Social Force Models (SFM)\cite{Helbing1995sfm} to define avoidance interactions between agents. 
However, these methods often suffer from generating smooth movement and limited diversity of human interactions.
Data-driven approaches have been proposed to model the more realistic behaviors with real-world data. 
For instance, SocNavBench\cite{biswas_socnavbench_2022} replays the pedestrian trajectory from the ETH/UCY dataset\cite{lerner2007eth, Pellegrini2009ucy}.
In order to control the movement of crowds in any simulation scenario without sacrificing the diversity and realism of human behavior, CPP\cite{Panayiotou2022cpp} uses RL to select optimal strategies from discrete behavior patterns. 
In addition, SocialGAIL\cite{Ling2024socialgali} employs generative adversarial imitation learning to imitate human social interactions and generate trajectories.
Therefore, NavIsaacLab adopts a diffusion model-based trajectory generation algorithm\cite{rempeluo2023tracepace} with multi-modal sampling, ensuring diversity and providing uncertainty quantification for human-aware navigation benchmark.

The generation of high-fidelity whole-body motion is particularly challenging due to its highly articulated nature, bio-mechanical constraints, and context-related coherence.
Current platforms\cite{tsoi_sean_2022, higueras_hunavsim_2023, vuong2024habicrowd} rely on pre-recorded animation to perform fixed patterns or behavior trees on human motion. 
However, the significant gap lies in the lack of consistency between trajectories and actions, especially the visible discontinuities during motion transitions, such as abrupt changes in gait speed or posture.
Recent advances in human avatar and human motion generation leverage learning-based methods, including deep generative models and RL to improve motion diversity and naturalness\cite{tevet2023human,yi2024generating}.
Physics-based methods\cite{peng2021amp, taheri2020grab} further enhance realism by addressing environmental interactions and enforcing physical constraints, enabling collision avoidance and terrain adaptation.
Inspired by AMP\cite{peng2021amp, rempeluo2023tracepace}, our work creates pedestrian agents with joint-level control through imitation learning. 
Leveraging Isaac Lab’s parallelization capabilities, we achieve real-time control of dense crowds with simulated physical feedback, balancing physical realism with computational efficiency.

\section{Overview of NavIsaacLab Platform}

\begin{figure*}[!t]
\centering
    \includegraphics[width=0.98\linewidth]{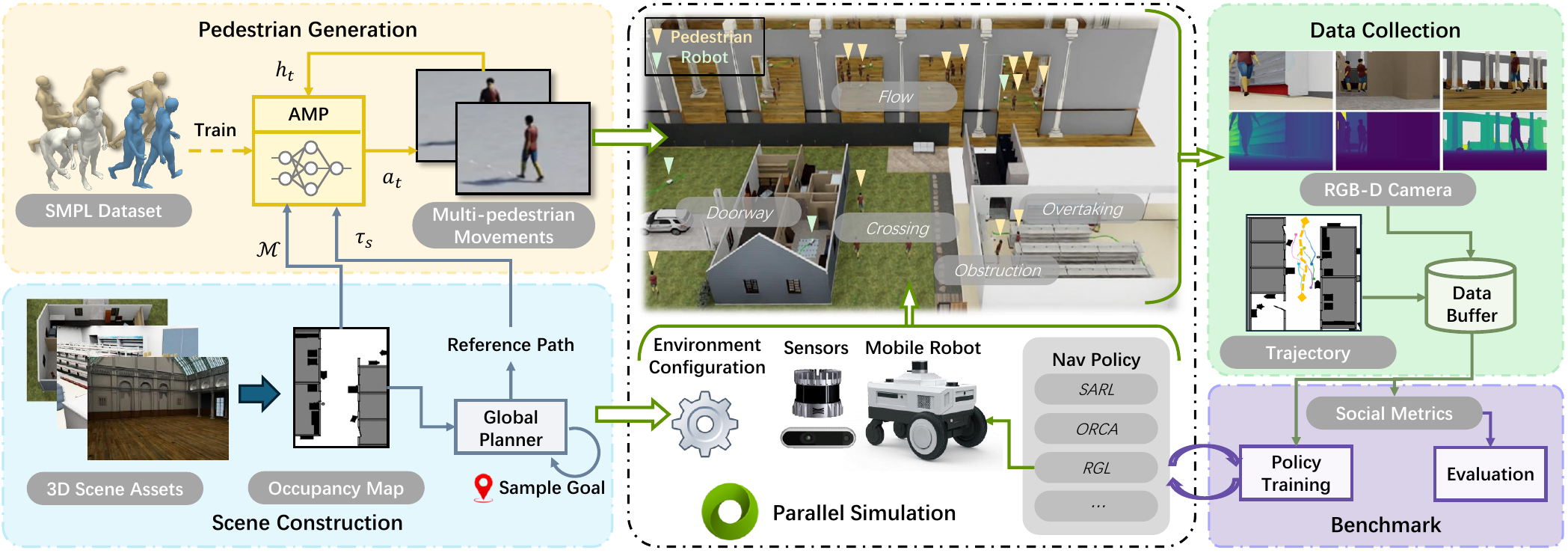}
    \caption{
    The framework of the proposed NavIsaacLab platform. 
    Prior to simulation, a data-driven pedestrian model is pre-trained, scene assets are curated, and the number and tasks of pedestrians within the simulation are stochastically configured. 
    During the simulation, the robot and pedestrians leverage the Isaac Lab framework to perform parallel operations across multiple scenes, acquiring real-time feedback. 
    Finally, multi-modal data is collected over multiple iterations, facilitating online policy learning or the comprehensive evaluation of existing navigation algorithms.
    }
    \label{fig:frame}

\end{figure*}


To support the training and evaluation of robot navigation policies in dynamic and complex environments, we introduce NavIsaacLab to generate high-fidelity interactive data at scale. 
Built on the IsaacLab\cite{mittal2025isaaclab} framework, NavIsaacLab extends GPU-accelerated simulation capabilities to everyday human-populated navigation scenarios.
IsaacLab inherits a kernel of the physics engine and realistic rendering, which provides a standardized framework that supports robot learning in massively parallelized environments.
On this foundation, we develop a seamless pipeline for generating navigation scenarios with natural pedestrian activities and a benchmarking suite for data collection and performance analysis of navigation policies.
As illustrated in Fig.~\ref{fig:frame}, the core functionalities of NavIsaacLab are organized into four modules:
(1) Scene Construction, which constructs static 3D scenes and defines global navigation paths;
(2) Pedestrian Generation, which synthesizes natural human behavior using data-driven priors;
(3) Parallel Simulation, the Isaac Lab simulation engine that synchronizes robots, pedestrians, and environments to support high-throughput rollout generation; and
(4) Benchmarking Suite, which manages multimodal sensing and provides standardized evaluation protocols grounded in human-aware navigation metrics.


\subsection{Scene Construction}
\label{sec:Scene}

The diversity and realism of simulated scenarios are essential for evaluating and improving the generalizability of navigation policies. 
NavIsaacLab leverages photorealistic 3D environments that incorporate geometric structure, semantic context, and visual texture to construct visually and physically grounded navigation spaces. 
All scene elements are defined in USD format and managed by the Scene Graph of Isaac Lab, where properties such as mesh geometry and material textures are explicitly defined for realistic rendering and accurate physical interaction. 
NavIsaacLab includes a predefined benchmark database of artist-crafted scenes derived from the OmniGibson Dataset\cite{li2022behavior}.
These scenes span a wide range of spatial scales and topological complexity to evaluate navigation policies under diverse spatial layouts.
A scene in NavIsaacLab consists of physically grounded architectural structures (e.g., walls and floors), semantically defined functional areas (e.g., kitchens, living rooms, or hallways), and context-appropriate household objects populate the scene to simulate real-world conditions.

For task execution, NavIsaacLab generates an occupancy grid from the static scene layout to define the traversable space for subsequent motion planning. Tasks are defined as start and end points with collision-free paths. 
We conduct repeatable experiments by controlling the sampling of robot initial states and navigation goals, while ensuring the diversity of dynamic navigation scenarios. 
Global reference paths are computed using an A* planner that guarantees path optimality and supports flexible waypoint intervals. 
The global planner uniformly compute paths for multiple agents and does not distinguish between pedestrians and robots.

\begin{figure*}[htbp]
\centering
    \includegraphics[width=0.88\linewidth]{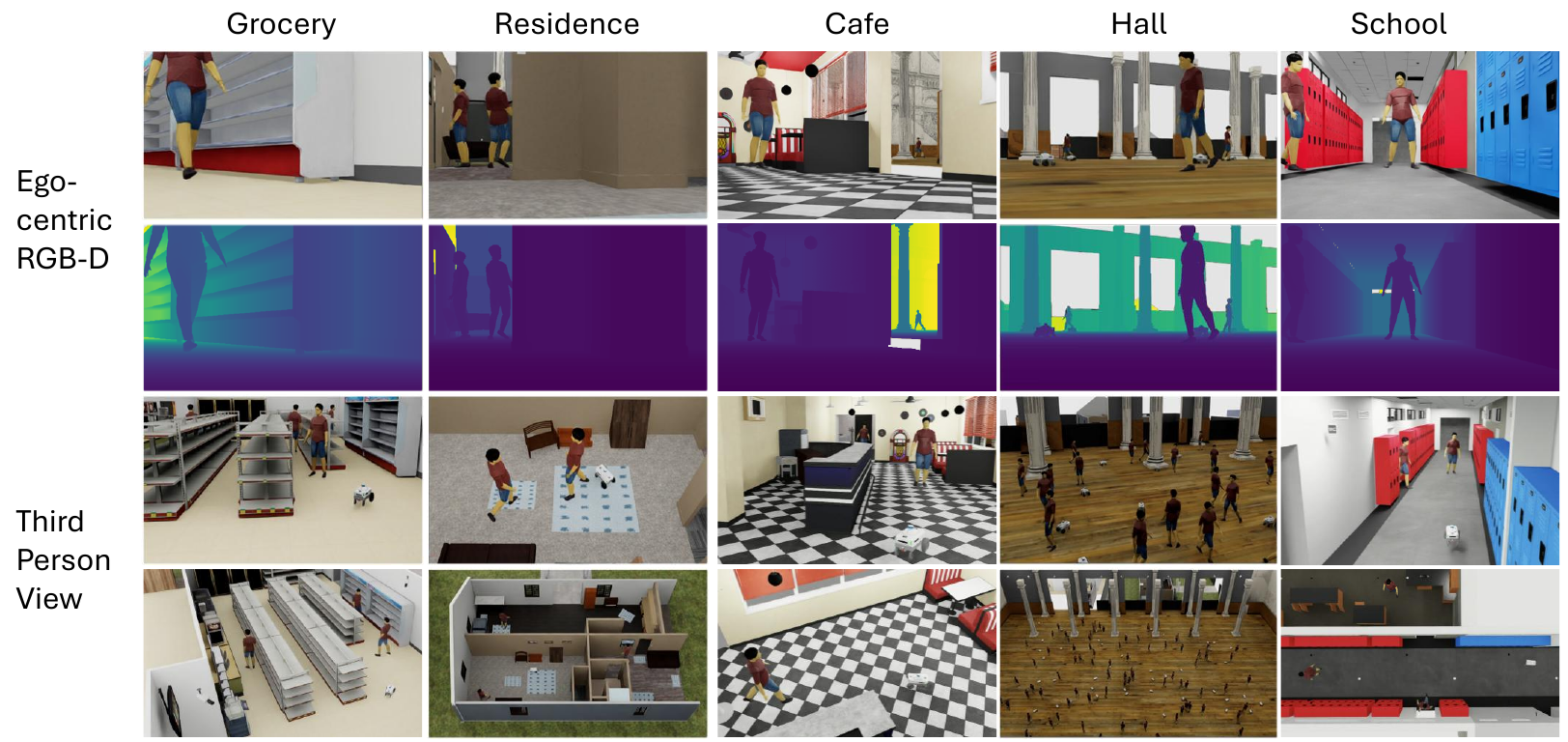}
    \caption{
    Rendering of several simulation scenes and real-time observation from the robot's perspective. The same robot (white) is adopted, and multiple robots simulated in parallel can be seen in the Hall scene.
    }
    \label{fig:vision}
\end{figure*}

\subsection{Pedestrian Generation}
Realistic pedestrian movements are essential for simulating dynamic real-world environments and enabling meaningful interaction among agents. 
NavIsaacLab aims to support flexible, natural, and diverse pedestrian behaviors—such as turning, running, avoidance, and reactive responses—beyond what scripted animations can provide, especially as the number of agents increases. 
To this end, NavIsaacLab introduces fully controllable and physically consistent virtual pedestrian agents together with a physics-based pedestrian controller. 
These agents are modeled as articulated bodies with actuated joints based on the Isaac Lab simulation engine, allowing the reproduction of natural whole-body kinematics\cite{rempeluo2023tracepace}.
Each pedestrian agent is configured following the SMPL model, which represents the human body through learned parameters of body shape coupled with the kinematic skeleton\cite{loper2015smpl}.
For total body control of imitating real-world pedestrian motion, we map the skeleton of the SMPL model to 24 actuated joints rooted in the pelvis and spanning major joints such as the hips, knees, and neck. 
A corresponding skinned mesh generated from the SMPL shape parameters provides surface geometry for photorealistic RGB-D rendering.

Pedestrian behavior is decomposed into the local trajectory generation at the agent level and the whole-body motion control at the joint level. 
First, given a sequence of waypoints in Euclidean space, the trajectory generation module computes a collision-free trajectory that accounts for multi-agent interactions such as cooperative avoidance. 
NavIsaacLab provides both traditional rule-based planners, such as SFM, and interfaces for deploying learning-based multi-agent trajectory generators on the GPU to support large-scale crowd scenarios.
Besides, whole-body motion control is implemented through adversarial motion learning, in which a pedestrian agent policy network is trained from real-world human motion data to track the reference trajectory and compute coherent joint-level actions. 
Due to the GPU-accelerated parallel computation, a large number of pedestrian agents can coexist and interact within a shared environment with the learned policies.

\subsection{Simulation and Benchmark of Navigation Policies}
The aforementioned scene database and pre-trained pedestrian motion model are integrated into the Isaac Lab workflow. 
Before the simulation begins, robots are instantiated through configuration files that define their base and sensor parameters, and continuous or discrete action spaces to match different navigation policies. 
While users may incorporate custom robot models, our experiments adopt the differential-drive mobile base Nova Carter as an exemplified platform.
Then, multiple scenes from the database can be simultaneously spawned into an extended simulation space to maximize sampling efficiency.
Finally, all robots and pedestrian agents are managed through vectorized simulation environments, enabling real-time control of multiple agents in parallel. 
Observations, actions, and rewards from each environment are wrapped as an OpenAI Gym interface\cite{openaigym} that supports batched inputs for the policy model of any agent.
Fig.~\ref{fig:vision} demonstrates 5 simulation scenarios along with the corresponding robot first-person views. 
Notably, observation data from robots can be collected synchronously, both within the same scenario and across different scenarios.

During simulation, the system captures rich perception data, such as RGB-D camera streams and agent states, such as trajectories. 
In training mode, these data support interactive rendering and real-time feedback, enabling continuous refinement of navigation policies. NavIsaacLab is compatible with Stable-Baselines3\cite{sb3}, a standard reinforcement learning library, facilitating the development of human-aware navigation strategies. 
In our experiments, we also demonstrate how a vision-based RL approach can effectively leverage the dynamic visual cues generated by simulated pedestrian agents.
In evaluation mode, the platform provides a comprehensive benchmarking suite that quantifies navigation performance using human-aware metrics related to safety, efficiency, and social compliance. 
The scalable combination of diverse scenes and realistic pedestrian simulations allows NavIsaacLab to generate a wide range of navigation situations, including narrow doorways, bidirectional crossings, and obstructions.
We report benchmark results for several representative navigation methods of systematic comparison across different settings.

\section{Data-driven Pedestrian Simulation}

A primary objective of NavIsaacLab is the enhancement of crowd simulation, while traditional methods, such as Optimal Reciprocal Collision Avoidance (ORCA) or SFM, struggle to capture the richness of real human motion.
In this section, we introduce the realization of a data-driven pedestrian simulation method capable of producing both realistic and controllable pedestrian behaviors, built upon the Trace and Pace frameworks\cite{rempeluo2023tracepace}.
NavIsaacLab develops a hierarchical pedestrian simulation algorithm that integrates multi-agent trajectory planning to generate collision-free pedestrian paths that satisfy user-defined goals, together with a whole-body controller that produces safe and natural locomotion for tracking these trajectories.

\subsection{Trajectory Generation}

We represent each pedestrian using a root state indicating planar motion, $s = [x, y, u, v, \theta]$,
where $\mathbf{p} = (x, y)$ denotes the position, $\theta$ means the heading angle, and $\mathbf{v} = (u, v)$ means the velocity vector.
The trajectory diffusion model operates on a fundamental principle: it learns to gradually denoise a trajectory from pure noise to an available pedestrian path. 
At each time step, the model outputs a future trajectory plan for a target agent conditioned on its past trajectory, including the past trajectories of all neighboring agents and the environmental context. 
Given past observations $\tau_h=[s_0,s_1,...,s_T]$, neighboring trajectories $\mathbf{N}$, and occupancy map $\mathcal{M}$, the model outputs future states $\tau_s = [\mathbf{s}_{t+1},...,\mathbf{s}_{t+T_f}]$.
Each step of this denoising process is conditioned on $C$:
\begin{equation}
p_\phi(\tau_{k-1} | \tau_k, C) := \mathcal{N}(\tau_{k-1}; \mu_\phi(\tau_k, k, C), \sigma_k^2\mathbf{I}),
\end{equation}
where $\phi$ are model parameters, $\sigma_k$ is from a fixed schedule, and $C=(\mathbf{H},\mathbf{N},\mathcal{M})$ is the conditioning context.
At every step it learns to predict the final clean trajectory $\tau_0$, where training supervises this prediction $\hat{\tau}_0$ using ground-truth future trajectories via a mean squared error (MSE) loss.  

Specifically, we adopt a U-Net diffusion architecture\cite{rempeluo2023tracepace}.
As the input of the denoising network, the noisy trajectory input $\tau_k$ is concatenated with the context embeddings that are extracted by the convolutional encoder of the map features from $\mathcal{M}$.
The input trajectory is processed by temporal convolutional blocks that progressively downsample and upsample the temporal sequence with skip connections. 
The model generates a temporally continuous future trajectory conditioned on the pedestrian’s past motion, neighboring pedestrian motions, and map context.
Given sparse global navigation waypoints from the navigation task, the diffusion module refines them into temporally continuous local target trajectories. 
Following the trajectory diffusion design in TRACE\cite{rempeluo2023tracepace}, one model inference produces a future trajectory segment over a 5-second horizon. 
The generated trajectory is then converted into local targets and tracked by the AMP-based pedestrian controller during subsequent steps.

\subsection{Controllable Human Motion Generation}

\begin{figure}[!t]
\centering
    \includegraphics[width=0.98\linewidth]{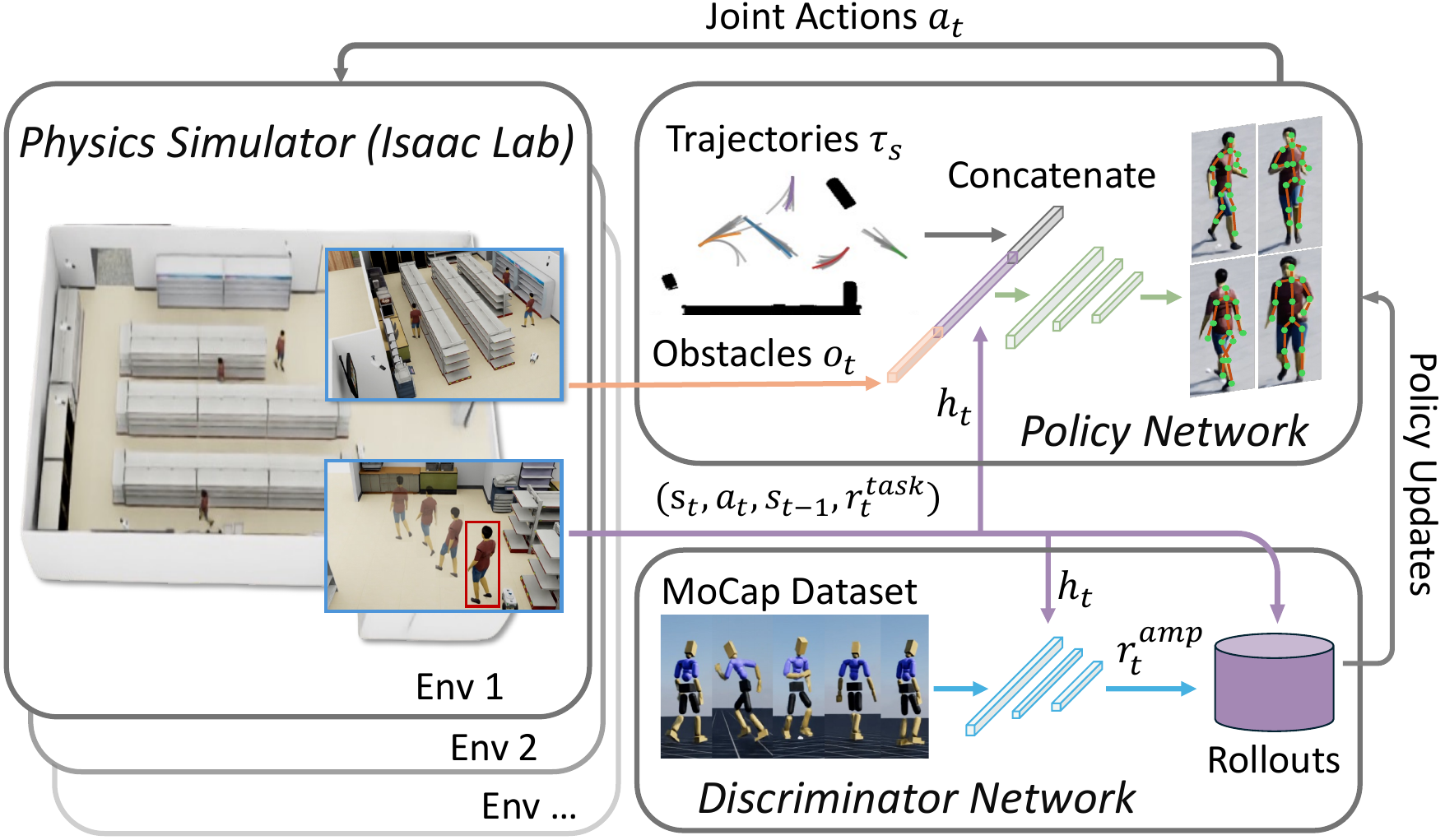}
    \caption{
    The algorithm framework of the whole-body pedestrian agent control based on Adversarial Motion Priors (AMP) to calculate joint-level realistic motion. 
    This methodology utilizes adversarial learning to guide the human agent to imitate motions from a real-world dataset while simultaneously tracking specific navigation goals. 
    The training is efficiently executed within the parallel simulation of the Isaac Lab framework.
    }
    \label{fig:amp}
\end{figure}

In order to achieve physics-based whole-body control of the pedestrian agent, we design a controller based on adversarial motion priors (AMP)\cite{peng2021amp} to enable pedestrians to follow 2D trajectories with realistic motions. 
As depicted in Fig.~\ref{fig:amp}, our framework aligns with the standard frame of goal-conditioned reinforcement learning. 
Here, we train the goal-conditioned policy $\pi$, whose objective is to control the agents to track the 2D target trajectories denoted by $\tau_s$. 
The objective of the policy is to maximize the expected discounted return $\mathbb{E}\left[\sum_{t = 1}^{T}\gamma^{t - 1}r_{t}\right]$, where $r_t$ represents the reward obtained at each timestep. 
As shown in Fig.~\ref{fig:amp}, our control policy $\pi(a_t|h_t, o_t, \tau_s)$ is conditioned on the state of the simulated agent $h_t$, obstacle features $o_t$, and 2D target trajectory $\tau_s$. 
Notably, $h_t$ can be calculated from $s_t$, where $s_t$ represents the agent's state at each timestep.

AMP\cite{peng2021amp} is employed to learn the optimal control policy $\pi^*$, which has the ability to not only follow the given 2D trajectory but also generate realistic pedestrian motions.
As shown in Fig.~\ref{fig:amp}, it uses a motion discriminator to compare the actions generated by the current policy $\pi$ with the realistic motion capture data. 
In each step, the discriminator is used to compute a motion style reward $r^{\text{amp}}_t$ that encourages the policy to optimize itself in the direction of generating realistic actions. 
Formally, the reward function can be expressed as:
\begin{align}
    r^{\text{amp}}_t &= \text{D}(h_{t-10:t}, a_t),
\end{align}
where D$(\cdot)$ represents the motion discriminator. 
In practice, we use a subset of the AMASS dataset\cite{AMASS} as the reference motions, such as walking, jogging, and looking around.

The policy input consists of a humanoid self-state and a task observation.
The task observation is encoded and combined with the self-state feature before being passed to the actor and critic networks. 
The actor outputs continuous joint-level control actions for the SMPL-based humanoid, while the critic estimates the value function for PPO training.
The policy network and the AMP discriminator are both implemented as a compact MLP-based policy.

The total reward is the combination of the AMP reward and a task-specific reward $r^{\text{task}}_t$ that guides trajectory tracking and physical feasibility: $r_t = r^{\text{amp}}_t + 0.5r^{\text{task}}_t.$
In addition to the AMP reward, there is another reward term $r^{\text{task}}_t$ related to the task, which can be calculated based on the current state $S$ of the agent. 
The task reward $r^{\text{task}}_t$ consists of three components:
\begin{align}
    r^{\text{task}}_t &= r^{\tau}_t + r^{e}_t + r^d_t,
\end{align}
where $r^{\tau}_t$, $r^{e}_t$, and $r^{d}_t$ correspond to the error of trajectory following, the energy of the agent's actions, and the correctness of the agent's motion direction.
In detail, they are formulated respectively as follows:
\begin{align}
    r^{\tau}_t &= \exp(-\alpha_{p}\delta_t) ,\\
    r^{e}_t &= -\alpha_e\sum_{i = 1}^{n}\Gamma^i_t \omega^{i}_t ,\\
    r^{d}_t &= 1-\exp(\frac{\min(0, u_t)}{\alpha_d}) .
\end{align}
It is noted that $r^{d}_t$ can prevent the agent from walking backward during training.
$\delta_t$ represents the sum of the squares of the distance differences between the position of the agent and the 2D target trajectory. 
Besides, $\Gamma^i_t$ and $\omega^i_t$ are the torque and velocity of the joint $i$ , and $u_{t}$ is the root velocity of the agent in the forward direction within the agent's ego-centric coordinate. 
Particularly, we set the parameters $\alpha_p = 2.0$, $\alpha_e = 0.0005$, and $\alpha_d = 1.8$ in training.

\subsection{Quantitative Evaluation of Pedestrian Simulation}

To train the AMP-based whole-body controller, we collect 2,848 motion clips from the AMASS dataset as reference motion styles. 
These clips include walking, running, turning, sidestepping, and transition motions, and are filtered according to their labels to focus on navigation-related locomotion. 
To assess the effectiveness of the learned controller in Isaac Lab, we use the learned AMP policy to track the reference motions in a closed-loop manner. 
During rollout, the simulated rigid-body states are synchronized with the reference sequence and compared frame by frame.
The learned AMP policy achieves a mean rigid-body position error of 0.218 m, a mean rigid-body rotation error of 0.343 rad, and a maximum rigid-body position error of 0.433 m. 
Under a stringent failure criterion where a sequence is regarded as failed once the per-step mean rigid-body position error exceeds 0.5 m, the controller successfully tracks 64.6\% of the evaluated locomotion clips. 
These results indicate that the AMP controller provides sufficient full-body tracking capability for navigation-oriented pedestrian simulation, while some challenging locomotion transitions and long-tail motions remain difficult to reproduce accurately.

\begin{table}[h]
\centering
\caption{Navigation reliability of the integrated pedestrian motion generation method.}
\label{tab:navigation_quality}
\resizebox{0.49\textwidth}{!}{%
\begin{tabular}{lccccc}
\toprule
Method & Success  $\uparrow$ & Fall  $\downarrow$ & Map Col.  $\downarrow$ & Agent Col. $\downarrow$ & Path Error$\downarrow$ \\
\midrule
SFM+AMP & 74.24 & 4.99 & 3.74 & 17.17 & 0.189 m \\
Diffusion+AMP & \textbf{88.67} & \textbf{1.12} & \textbf{2.40} & \textbf{7.81} & \textbf{0.153 m } \\
\bottomrule
\end{tabular}
}
\end{table}

To further validate the fidelity of the proposed pedestrian simulation module, we evaluate the pedestrian motion generation method by navigation reliability, and motion diversity. 
These experiments are designed to verify whether the proposed method can generate controllable full-body pedestrian behaviors, rather than only visually plausible animations and can be reliably deployed in different navigation scenarios. 
Besides, we compare the proposed Diffusion+AMP pipeline with an SFM+AMP variant, where the same AMP-based articulated body controller is retained but the diffusion-based trajectory generation module is replaced by an SFM-based target generator. 
Therefore, this ablation isolates the contribution of the high-level trajectory generation module under the same low-level whole-body controller. 
The evaluation is conducted in 8 scenes with 3--15 pedestrians per scene, and each scene is sampled for 200 s. 
We report success rate, fall rate, map collision rate (Map Col.), agent collision rate (Agent Col.), and the path error.
As shown in Table~\ref{tab:navigation_quality}, replacing SFM with the diffusion-based trajectory generator consistently improves navigation reliability. 
These results indicate that the diffusion module produces smoother and more feasible high-level targets for the articulated pedestrian controller, thereby improving both physical stability and multi-agent interaction quality.

\begin{table*}[h]
\centering
\caption{Motion diversity and AMASS-likeness compared with IsaacSim built-in GoTo animation.}
\label{tab:motion_diversity}
\begin{tabular}{lcccccccc}
\toprule
Method & Coverage (K=50)$\uparrow$ & H (K=50)$\uparrow$ & Dist.(K=50)$\downarrow$ & Coverage (K=100)$\uparrow$ & H (K=100)$\uparrow$ & Dist. (K=100)$\downarrow$  & MMD $\downarrow$ \\
\midrule
AMASS reference & 1.00 & 0.888 & 0.190 & 1.00 & 0.890 & 0.171 & 0.023 \\
IsaacSim Animated People & 0.34 & 0.641 & 0.229 & 0.35 & 0.648 & 0.226 & 0.771 \\
\textbf{Ours} & 0.92 & 0.800 & 0.305 & 0.90 & 0.821 & 0.285 & 0.327 \\
\bottomrule
\end{tabular}

\end{table*}

Finally, we evaluate whether the proposed method generates more diverse full-body pedestrian motions than the built-in animation-based pedestrians in Isaac Sim. 
All motion sequences from different system are converted into a unified 15-keypoint representation, including pelvis, hips, knees, feet, shoulders, elbows, and wrists. 
The poses are resampled to 30 Hz, root-centered, body-heading aligned, and scale-normalized. 
We compute AMASS-cluster coverage and normalized pose entropy to measure pose diversity, and Motion-MMD (Maximum Mean Discrepancy) to measure distributional similarity to AMASS locomotion data.
Table~\ref{tab:motion_diversity} shows that our method covers substantially more AMASS pose modes than the built-in IsaacSim Animated People baseline. 
The normalized pose entropy also increases from 0.641 to 0.800 for $K=50$ and from 0.648 to 0.821 for $K=100$, indicating that the proposed pedestrians simulation method occupy a broader region of the human pose space. 
In addition, Motion-MMD is reduced from 0.771 to 0.327, suggesting that the generated motion distribution is closer to real locomotion data from AMASS. 
These results demonstrate that the proposed pedestrian motion generation method improves both the diversity and AMASS-likeness of full-body pedestrian motion compared with animation replay.

\section{Demonstration of Training Vision-based Human-aware Navigation Policy}

\begin{figure}[!t]
\centering
    \includegraphics[width=0.98\linewidth]{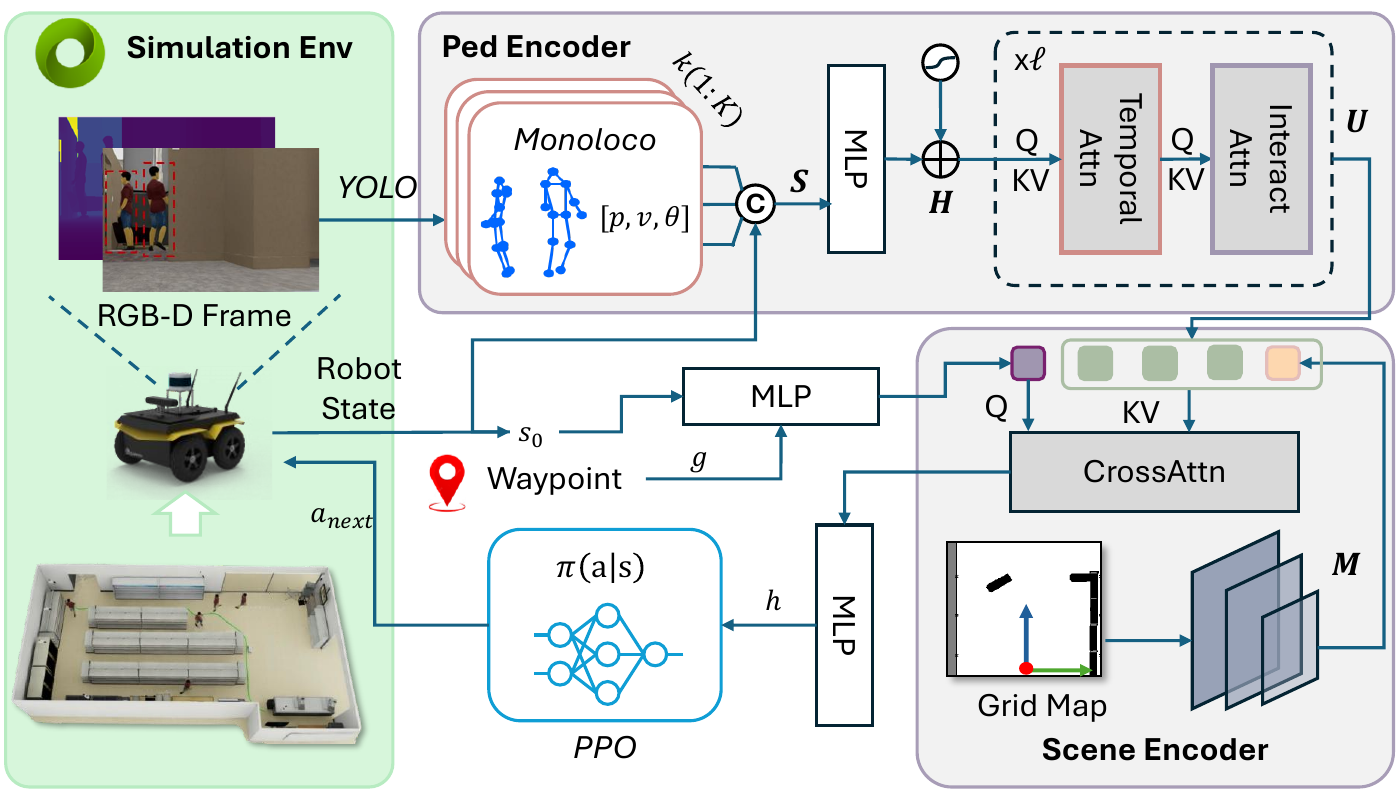}
    \caption{
    The proposed policy network for RL-based human-aware navigation. 
    A crucial component involves encoding the human body pose observed from ego-centric vision to infer the human movement intention based on the difference between their pose orientation and velocity. 
    Furthermore, we construct the RL state encoding network using a Transformer-based architecture to capture the interaction of the robot and surrounding pedestrians under the environmental context. 
    Both visual and physical feedback streams are acquired in real-time through the NavIsaacLab simulation framework.    
    }
    \label{fig:gaf}

\end{figure}

The NavIsaacLab simulation environment offers diverse, realistic scenarios and high training efficiency. 
Leveraging these capabilities, we propose a baseline policy to achieve safe and efficient robot navigation in dense crowds. 
Our approach captures pedestrian motion from ego-centric vision and embeds this information within a DRL framework. 
As shown in Fig.~\ref{fig:gaf}, the full framework consists of three primary modules: (1) Perception and feature extraction of pedestrians, (2) A Transformer model for feature fusion, and (3) an RL-based policy for human-aware navigation.

The objective is to learn a navigation policy $\pi(a_t \mid \mathcal{O}_t, g),$
where $\mathcal{O}_t$ denotes multi-agent historical observations and $g$ denotes the navigation goal.
All observations are expressed in the robot’s local coordinate frame at the current time step $T$, such that historical trajectories are aligned with the robot’s heading.
At each time step, the robot and surrounding agents (K agents) are considered.
Their state is uniformly represented by planar position, velocity, and orientation, denoted as $s = [x, y, u, v, \theta]$. 
We adopt the past $T=8$ steps trajectory.

\subsection{Pedestrians Observation Encoder}


Conventional DRL navigation policies often overlook crucial visual cues of a pedestrian's motion, focusing instead solely on human position and velocity. 
To address this limitation, we additionally estimate pedestrian body orientation from visual observations.
Incorporating orientation information enables the network to more accurately infer motion inertia and future movement intentions.
This module performs feature processing in two stages: individual pedestrian state estimation and the Self-Attention Mechanism\cite{vaswani2017attention}.

\textbf{Pedestrian State Estimation.}
In our approach, the key is to estimate the essential body orientation component from ego-centric vision.
We adopt Monoloco++\cite{monoloco} for accurate 3D human detection and localization from single RGB images, primarily utilizing the precise sensing of body orientation $\theta$ for every visible pedestrian.
First, a 2D pose estimation technique extracts human keypoints through an off-the-shelf pose detector network. 
Then, a feed-forward network regresses the human location in ego-centric spherical coordinates to compute the relative human orientation, leveraging normalized 2D joints without explicit supervision on depth ambiguity. 
Furthermore, the use of the depth camera and existing human tracking methods\cite{yolo11_ultralytics} allows us to estimate the pedestrian's position across sequential images and calculate the position difference as the velocity $\mathbf{v}$.
The full observation $\mathbf{S}=\{\hat{s}_{k,t}\} \in \mathbb{R}^{K \times T \times F}$ consists of all the agents' extended states, where $(\cos\theta_{k,t}, \sin\theta_{k,t})$ replace the angle $\theta$ and the agent radius $\rho_k$ is included.
The robot is assigned a fixed index of 0.

\textbf{Factorized Attention Block.}
Our model follows a factorized spatio-temporal Transformer architecture that decomposes temporal dynamics and agent interactions into separate attention operations. 
The network consists of three sequential spatio-temporal attention blocks.
Each agent feature vector is first projected into a $D$-dimensional latent space using a shared multilayer perceptron:
\begin{equation}
\mathbf{h}_{k,t} = \mathrm{MLP}(\hat{s}_{k,t}) \in \mathbb{R}^{D}.
\end{equation}
All the states $\mathbf{S}$ are mapped as original features $\mathbf{H} \in \mathbb{R}^{K \times T \times D}.$
To encode the temporal order, a learnable temporal positional embedding is added to the feature.

The block begins with temporal self-attention applied independently to each agent. 
For a fixed agent $k$, the history sequence $\{\mathbf{h}_{k,t}\}_{t=1}^T$ is processed using multi-head self-attention (MHSA) along the temporal axis:
\begin{equation}
\widetilde{\mathbf{H}}^{\ell}_{k} = 
\mathrm{MHSA}\big(\mathbf{h}_{k,1:T}\big).
\end{equation}
This operation captures temporal motion patterns while preserving agent identity. 
A strict causal mask is applied such that only history trajectory tokens are included.
Following temporal modeling, interactions among agents are captured at each time step using agent-wise attention:
\begin{equation}
\mathbf{H}_t^{\ell} = \mathrm{MHSA}(\widetilde{\mathbf{h}}_{0:K,t}^{\ell}).
\end{equation}
This design explicitly models social interactions among neighboring agents.
Thus, $\mathbf{H}^{\ell} \in \mathbb{R}^{K \times T \times D}$ is obtained by combining the updated tokens, yielding a refined spatio-temporal representation for the interactions between agents.

\subsection{Scene Context Encoder}

To incorporate static environmental constraints, we apply a single map interaction layer after spatio-temporal encoding. 
After stacking $\ell=3$ Factorized Spatio-Temporal Encoder blocks, we obtain the interaction-aware spatio-temporal representations of all agents.
We construct neighboring agents' interaction features by extracting the last-step tokens:
$\mathbf{U} = \mathbf{H}^{\ell}_{1:K,T} $.
The occupancy map is encoded as a geometric vector where the map feature is calculated by a CNN encoder as $\mathbf{M} \in \mathbb{R}^{D}$.
We construct a scene-level feature by concatenating the interaction feature and the map feature:
\begin{equation}
\mathbf{K}_{\text{scene}} = \mathbf{V}_{\text{scene}} =
\mathrm{concat}(\mathbf{U}, \mathbf{M}).
\end{equation}


The future goal position $g = (x_g, y_g)$ is given in the robot coordinate frame and expanded as $\mathbf{e}_g = [x_g,\; y_g,\; d_g,\; \sin \theta_g,\; \cos \theta_g]$, where $d_g$ and $\theta_g$ denote the distance and relative direction of the goal, respectively.
Combined with the robot's current state, the goal is encoded into a query embedding using a multi-layer perceptron:
\begin{equation}
\mathbf{q}_{\text{g}} = \mathrm{MLP}([\mathbf{s}_{0,T}; \mathbf{e}])
\end{equation}
where $d_g$ and $\theta_g$ denote the distance and direction angle of the goal, respectively.

We employ cross-attention to retrieve goal-relevant interaction context from the scene memory:
\begin{equation}
\mathbf{z} = \mathrm{CrossAttn}
\big( \mathbf{q}_g,\; \mathbf{K}_{\text{scene}} ,\; \mathbf{V}_{\text{scene}}
\big)
\end{equation}
This operation allows the goal query to selectively attend to surrounding agents and map elements that are most relevant for achieving the target objective.
To emphasize goal-directed intent when human-robot interaction is weak, we apply a gated update\cite{hivt} for attention calculation:
\begin{equation}
\alpha = \sigma\!\left(W_\alpha [\mathbf{z}; \mathbf{q}]\right), \quad
\mathbf{h} = \alpha \odot \mathbf{q} + (1-\alpha) \odot \mathbf{z},
\end{equation}
yielding the full state representation $\mathbf{h} \in \mathbb{R}^D$. 
The operator $\sigma(\cdot)$ means the sigmoid function, and $\odot$ denotes element-wise product.

\subsection{DRL-based Navigation Policy}
The final stage synthesizes the robot's goal-directed state with the crowd's encoded social context to form the final Markov Decision Process (MDP) state.
The state representation $\mathbf{h}$ is passed into the PPO\cite{ppo} Policy Network $\pi(a|s)$, to compute the robot's action $a=(v, \omega)$, ensuring that the decision is informed by both its own goals and a robust, attention-weighted understanding of the surroundings.

To enable collision-free navigation in dynamic and complex environments, we design a multi-objective reward function: $r_t=r_t^\text{nav}+r_t^{\omega}$.
In particular, $r_t^\text{nav}$ encourages the robot to move toward the goal while maintaining a safe distance from pedestrians and obstacles:
\begin{align}
r_t^\text{nav} = \begin{cases}
20, & \mathrm{if}\ d_t^g< \rho \\
-20, & \mathrm{else\ if}\ d_t^o< 0 \\
0.5(d_t^o - 0.9), & \mathrm{else\ if}\  0\leq d_t^o < 0.9 \\
3.2(d_{t-1}^g - d_{t}^g), & \mathrm{otherwise},
\end{cases}
\end{align}
where $\rho$ is the radius of the robot, $d^g_t$ is the distance between the robot and its goal at time $t$, and $d^o_t$ is the minimum distance between the robot and any pedestrians or obstacle at time $t$.
Additionally, $r_t^{\omega}$ encourages smooth navigation by penalizing sharp turns: 
\begin{align}
r_t^{\omega} = \begin{cases}
-0.1|\omega_{t}|, & \mathrm{if}\ |\omega_{t} |> 1.0 \\
0, & \mathrm{otherwise},
\end{cases}
\end{align}
where $\omega_{t}$ denotes the robot’s angular velocity at time step $t$.

\subsection{Effects of Parallel Training}
\label{sec:Parallelism}
In this section, we study the effects of the number of parallel robots on the navigation performance of the policy. 
In order to improve the performance in different scenarios, based on the diverse scenarios provided by our simulator, we have designed a parallel training environment that incorporates multiple scenarios. 
Specifically, it includes: a station with a dense and crowded crowd, a complex home with multiple rooms, and a regular-shaped grocery store shopping environment with multiple aisles.

In the experiment, we employ the final reward metric to evaluate the performance of the policy. 
It should be noted that the final reward is defined as the average value of the rewards obtained in the last 10 episodes throughout the entire training. 
We uniformly implement the navigation policy mentioned above to all the simulated robots and synchronize the network weight across them.
All the results are obtained under the parameter settings of $n_{steps}$ = 32 and $n_{epochs}$ = 6.
We then conduct experiments in which we increase the number of robots while keeping a constant batch size. 
It should be noted that the case where the number of robots is 1 can be regarded as a Gazebo simulation. 

\begin{figure*}[htbp]
\centering
    \subfloat[]
    {\includegraphics[width=0.48\linewidth]{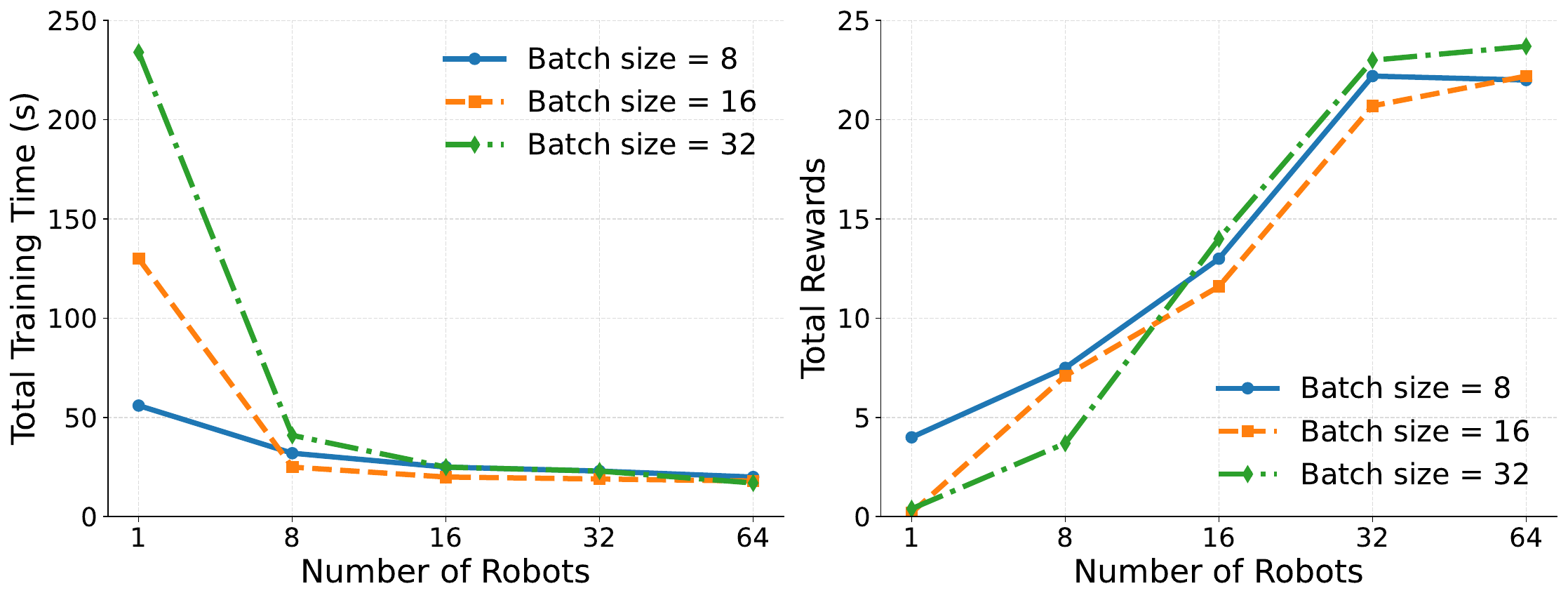}}
    \subfloat[]    
    {\includegraphics[width=0.48\linewidth]{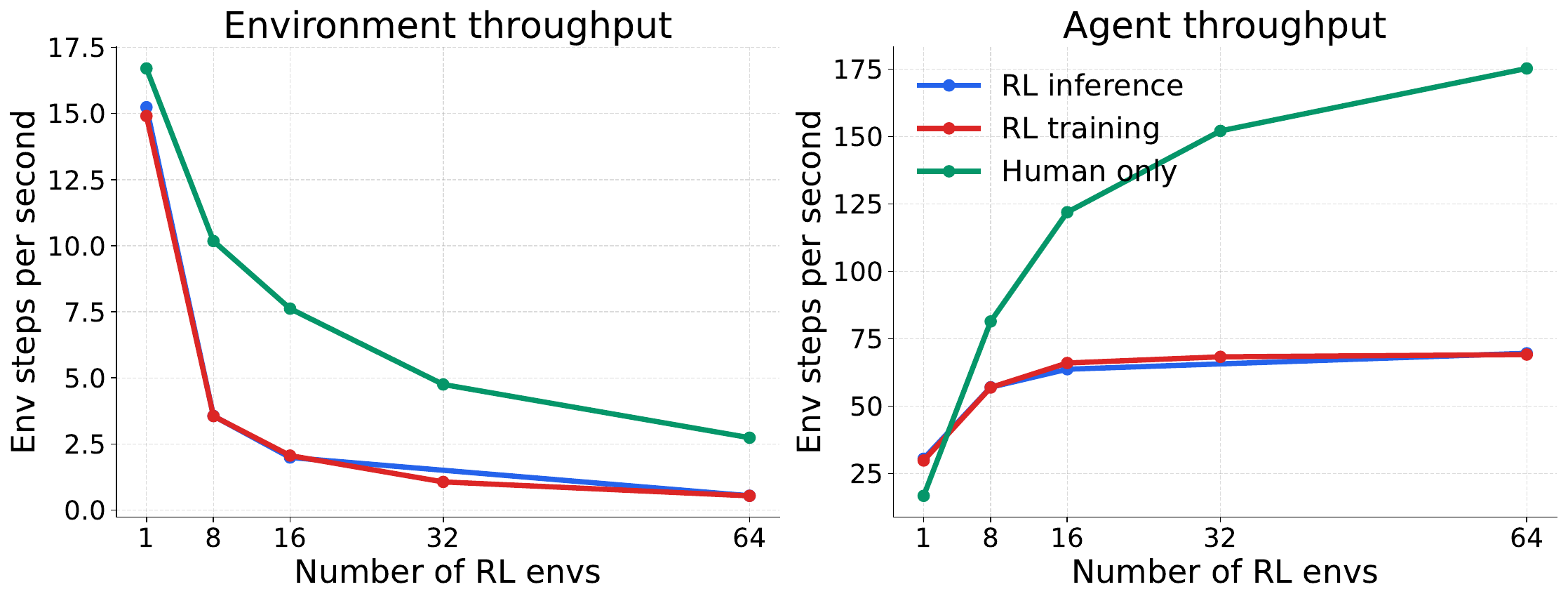}}
    \caption{
    Impact of parallel simulation of NavIsaacLab for human-aware policy training. 
    Different batches of state inputs in RL are compared for the demonstration of the computation efficiency in rollout sampling. 
    (a) The left figure shows The total training time after 5000 policy updates, and the right one shows the total rewards after 16000 simulator steps.
    (b) Runtime scaling of the pedestrian and mobile robot simulation. 
    The left figure shows environment throughput, and the right one shows aggregate agent throughput. 
    The human-only curve isolates the scalability of articulated pedestrian control. 
    The RL inference and RL training curves evaluate the complete navigation setting, where each additional parallel environment adds both one robot and one human pedestrian.
    }
    \label{fig:parrel}

\end{figure*}

In the first experiment, each time we record the total training time after 5000 policy updates to verify whether our parallel environment could accelerate the training process, as shown in Fig.~\ref{fig:parrel}(a). 
As expected, as the number of robots increases, the total training time continuously decreases. 
For online reinforcement learning algorithms like PPO, it updates the policy every time it has collected data for $n_\text{steps}$. 
As the number of robots increases, the amount of data collected each time will significantly increase, which can be used for more frequent policy updates. 
Since the environments are simulated in parallel, the process of data collection will not significantly prolong with the increase in the number of robots. 
Therefore, increasing the number of robots can accelerate the speed of policy updates. 
In addition, using a larger batch size generally takes more time, because the amount of data processed in each policy update becomes larger. 

In the second experiment, we record the final reward after the simulator runs for 16000 steps to verify whether the parallel environment can enhance the performance of the policy, as shown in Fig.~\ref{fig:parrel}(a). 
Just as expected, the final reward increases as the number of robots increases. 
As the number of robots increases, the simulator can simultaneously simulate a variety of different situations, obtaining more abundant data for policy updates. Therefore, increasing the number of robots can enhance the performance of the policy. In addition, a larger batch size generally has the potential to enhance the performance of the policy, because the amount of data utilized during each policy update increases. 
It should be noted that, as shown in the case where $n_\text{robots}$ = 8, it is not always the case that a larger batch size will definitely lead to better performance. This is because the optimization process of reinforcement learning itself progresses in an upward trend with fluctuations.

Finally, we test the runtime efficiency of the simulator under different levels of parallelization to further characterize the computational cost of NavIsaacLab.
We measure two throughput metrics: environment steps per second and agent steps per second. 
All experiments are conducted on a workstation with an NVIDIA RTX 4090 GPU, an Intel i9-12900K CPU, and 64 GB RAM. 
The former measures the wall-clock speed of the vectorized simulation loop, while the latter measures the aggregate number of simulated agent transitions processed per second.

As shown in Fig.~\ref{fig:parrel}(b), the pedestrian controller scales effectively under batched execution. 
In the human-only setting, where only the SMPL pedestrian simulation and AMP policy inference are retained, the agent throughput increases from 16.7 agent steps/s with 1 pedestrian to 175.3 agent steps/s with 64 pedestrians. 
This result indicates that the AMP-based whole-body controller is not the dominant computational bottleneck by itself. 
Therefore, introducing joint-level pedestrian control increases motion fidelity without making the pedestrian simulation prohibitively slow.

We further evaluate the complete robot--crowd navigation setting, where each parallel environment contains one mobile robot and one articulated pedestrian. 
In both RL inference and training settings, the aggregate throughput improves consistently as the number of parallel environments increases, demonstrating that the simulator benefits from vectorized execution even under coupled robot--crowd interaction. 
However, the scaling becomes sublinear when the number of environments is further increased, suggesting that the full robot--crowd simulation loop begins to saturate. 
Since the human-only setting scales more effectively, the main bottleneck is not the AMP controller alone, but the coupled simulation pipeline, including robot--pedestrian interaction, observation construction, collision checking, reset handling, and physics synchronization.

\section{Sim-to-Real Validation}




\begin{figure*}[!t]
\centering

    \includegraphics[width=0.98\linewidth]{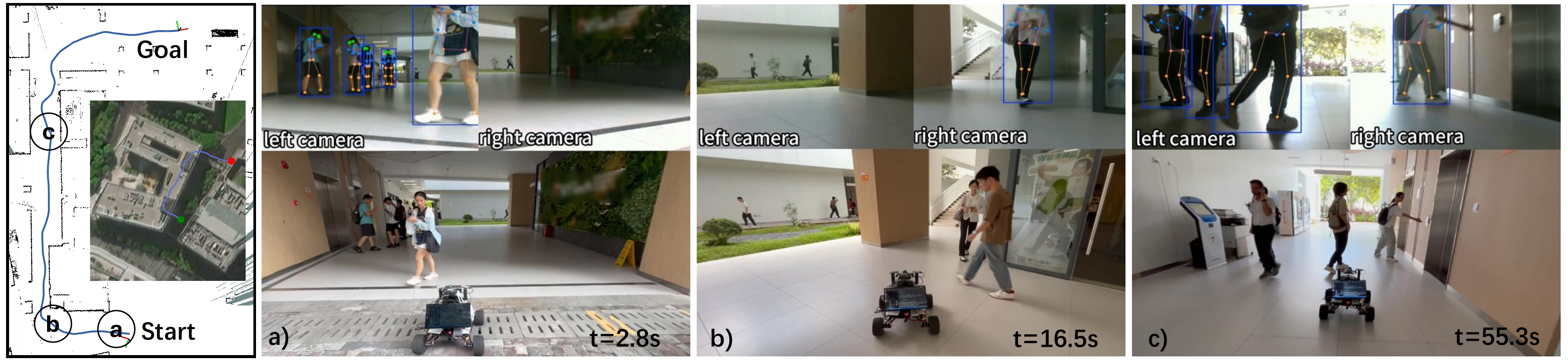}
    \caption{
    Snapshots and visualizations of the proposed method operating in the school corridor. 
    The subfigure on the left illustrates the map and the complete trajectory. 
    The marks represent the moments when the corresponding subfigures are shown.
    Each of them shows the robot’s camera frames and the corresponding real-world scene.
    }
    \label{fig:exp_school}
\end{figure*}

\begin{figure*}[!t]
\centering
   
    \includegraphics[width=0.98\linewidth]{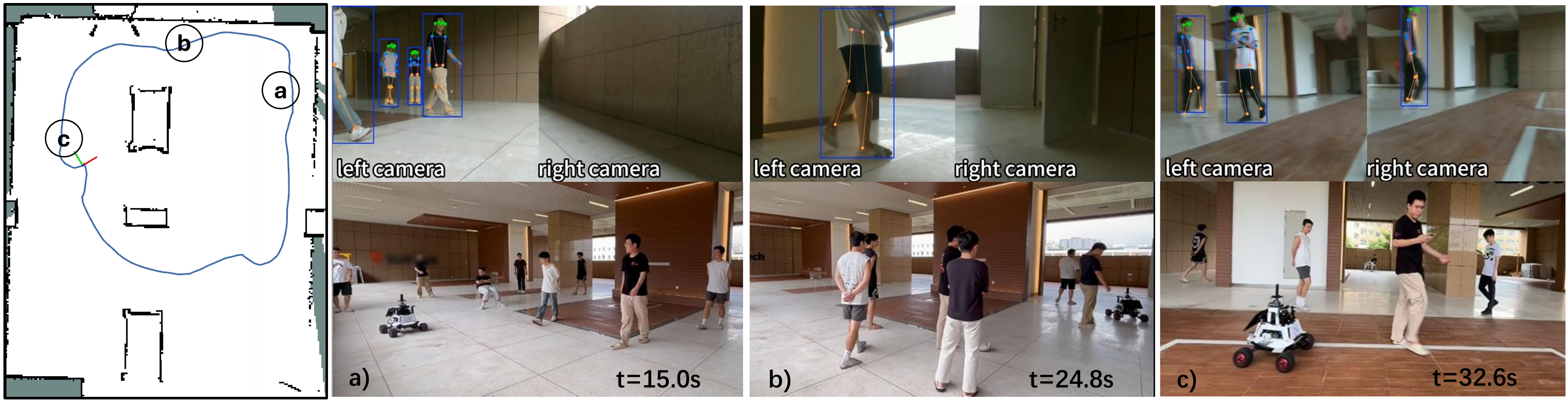}
    \caption{
    Snapshots and visualizations of the proposed method operating in a crowded lobby. 
    The subfigure on the left illustrates the map and the complete trajectory. 
    The robot executed a closed circular path back to the start position.
    The marks represent the moments when the corresponding subfigures are shown.
    Each of them shows the robot’s camera frames and the corresponding real-world scene.
    }
    \label{fig:exp_eng}

\end{figure*}

Beyond the simulated experiments, we conducted real-world experiments in complex environments to evaluate the performance of sim-to-real transfer. 
The algorithm, trained entirely in simulation, was deployed directly on the physical robot without tuning, while maintaining consistency in the robot’s intrinsic parameters with those in the simulation. 
The deployment platform consists of a Scout Mini robot equipped with two Intel RealSense D435i RGB-D cameras, each covering a 60-degree field of view on the left and right sides for local perception, while the global localization module leverages FAST-LIO\cite{fastlio} integrated with a Mid-360 LiDAR for real-time global localization of the robot.

Two experimental scenarios were implemented: one in an indoor lobby and another in a semi-outdoor school corridor. 
Key results are visualized in Fig.~\ref{fig:exp_school} and Fig.~\ref{fig:exp_eng}. 
In each figure, the left subfigure presents the occupancy grid map of the environment and the recorded robot’s actual trajectory, with subplots highlighting critical interaction snapshots. 
A complete demonstration of these experiments is available in the accompanying video.

In the first experiment, the task was navigation from an outdoor area into a teaching building, traversing a busy corridor. 
As illustrated in Fig.~\ref{fig:exp_school}(a), the robot encountered a pedestrian approaching from the opposite direction. 
By inferring the pedestrian’s intention to move leftward, the robot proactively adjusted its heading slightly to the left, creating sufficient clearance for the pedestrian to proceed unimpeded. This maneuver ensured smooth navigation while avoiding interference with another approaching pedestrian on the opposite side.

The second experiment, conducted in a crowded indoor lobby, tested the robot’s collision avoidance capabilities. 
The robot executed a closed circular path back to the start position.
As depicted in Fig.~\ref{fig:exp_eng}(a), the robot faced a constrained space caused by two approaching pedestrians and a static obstacle on its right. 
The system rapidly identified an open path on the front-left side and executed a timely steering maneuver to exit the confined area. 
This response minimized disruption to the nearby pedestrians while maintaining navigation efficiency.

\section{Conclusion}
In this work, we introduced NavIsaacLab, a comprehensive simulation framework designed to advance research in human-aware robot navigation. As robots are increasingly deployed in complex, dynamic environments shared with pedestrians, the need for high-fidelity, scalable, and socially realistic simulation tools has become more evident. Existing benchmarks often fall short in capturing multimodal human behaviors, providing diverse scene structures, or supporting parallel simulation at scale. NavIsaacLab addresses these limitations by integrating photorealistic rendering, physics-based scene and character modeling, data-driven trajectory generation, and a physically coherent whole-body pedestrian controller. Together, these components enable the creation of rich interactive environments that better reflect the challenges of real-world human-robot coexistence.

Beyond platform development, we established a unified benchmark that evaluates social navigation from multiple dimensions, including safety, efficiency, and social compliance. The proposed baseline policy demonstrates how NavIsaacLab can support the training of vision-based reinforcement learning strategies that leverage dynamic visual cues and cross-agent interactions. Moreover, our sim-to-real experiments confirm that the simulation fidelity and behavioral consistency provided by the platform enable direct policy transfer to real robots without fine-tuning, highlighting NavIsaacLab’s potential as a reliable tool for practical deployment.

To further enhance the flexibility and applicability of NavIsaacLab in the future, several key directions will be explored: (1) Supporting customizable pedestrian appearances (e.g., clothing, body types), (2) Expanding compatibility with diverse robot bases such as quadrupedal and humanoid, (3) Introducing more test scenarios, and (4) Developing intuitive task-definition interfaces via natural language.




\bibliographystyle{ieeetr}

\bibliography{ref.bib}


 





\end{document}